\documentclass[11pt,a4paper]{article}
\usepackage[hyperref]{naaclhlt2018}
\usepackage{times}
\usepackage{latexsym}

\usepackage{url}

\usepackage{extarrows}
\usepackage{amssymb,amsfonts}
\usepackage{latexsym}
\usepackage{multirow}
\usepackage{arydshln}
\usepackage{amsmath}
\usepackage{rotating}
\usepackage{color}
\usepackage{subfigure}
\usepackage{enumitem}

\newcommand{\ignore}[1]{}

\aclfinalcopy 

\def\eqref#1{Eq \ref{eqn:#1}}

\long\def\eat#1{\ignorespaces}

\title{Term Definitions Help Hypernymy Detection}

\author{Wenpeng Yin \rm{and} \textbf{Dan Roth}\\
	    University of Pennsylvania\\
	    {\tt \{wenpeng,danroth\}@seas.upenn.edu}}

\date{}

\newcounter{notecounter}
\newcommand{\enotesoff}{\long\gdef\enote##1##2{}}
\newcommand{\enoteson}{\long\gdef\enote##1##2{{
\stepcounter{notecounter}
\large\bf
\hspace{1cm}\arabic{notecounter} $<<<$ ##1: ##2
$>>>$\hspace{1cm}}}}
\enoteson
\enotesoff

\newcommand{\modelname}{\textsc{HyperDef}}

\begin{document}
\maketitle
\begin{abstract}
Existing methods of hypernymy detection mainly rely on statistics over a big corpus, either mining some co-occurring patterns like ``animals such as cats'' or embedding words of interest into context-aware vectors. These approaches are therefore limited by the availability of a large enough corpus that can cover all terms of interest and provide sufficient contextual information to represent their meaning.
In this work, we propose  a new paradigm, \modelname, for \textbf{hyper}nymy detection -- expressing  word meaning by encoding word \textbf{def}initions, along with context driven representation. This has two main benefits: (i) Definitional sentences express (sense-specific) corpus-independent meanings of words, hence definition-driven approaches enable strong generalization -- once trained, the model is expected to work well in open-domain testbeds; (ii) Global context from a large corpus and  definitions provide complementary information for words. Consequently, our model, \modelname, once trained on task-agnostic data, gets state-of-the-art results in multiple benchmarks\footnote{\url{cogcomp.org/page/publication_view/836}}.

\end{abstract}


\section{Introduction}\label{sec:intro}


Language understanding applications like textual entailment \cite{DRSZ13}, question answering \cite{DBLPSaxenaSKS07} and relation extraction \cite{DBLPMintzBSJ09}, benefit from the identification of lexical entailment relations. Lexical inference encompasses several semantic relations, with hypernymy being one of the prevalent \cite{DBLPRollerEB14,DBLPShwartzGD16}, an \emph{i.e., ``Is-A'' relation that holds for a pair of terms\footnote{This paper uses ``term'' to refer to any words or phrases.} ($x$, $y$) for specific terms' senses}.

Two families of approaches have been studied for identifying term hypernymy. (i) \emph{Pattern matching} exploits patterns such as ``animals such as cats'' to indicate a hypernymy relation from ``cat'' to 	``animal'' \cite{DBLPHearst92,DBLPSnowJN04}. However, it requires the co-occurrence  of the two terms in the same sentence, which limits the recall of this method; (ii) \emph{Term representation learning} depends on a vector embedding of each term, where each entry in the vector expresses an explicit context feature \cite{DBLParoniBDS12,DBLPRollerE16,DBLPSantusSS17} or a latent semantic \cite{DBLPFuGQCWL14,DBLP06371,DBLPGlavasP17}. 

Both approaches hinge on acquiring context-aware term meaning in a large corpus.  The generalization of these corpus-based representation learning paradigms, however, is limited due to the  domain specificity of the training data. For example, an IT corpus hardly mentions ``apple'' as a fruit.  Furthermore, the surrounding context of a term may not convey subtle differences in term meaning -- ``he" and ``she" have highly similar context that may not reveal the important difference between them. Moreover, rare words are poorly expressed by their sparse global context and, more generally, these methods would not generalize to the low resource language setting. 


Humans can easily determine the hypernymy relation between terms  even for words they have not been exposed to a lot, given a definition of it in terms of other words. For example, one can imagine a ``teaching'' scenario that consists of {\em defining} a term, potentially followed by a few examples of the term usage in text.


\ignore{We attribute this to the fact that human beings understand the meaning of those terms, which may be \emph{beyond the expressivity of distributional context words}. Then, how to make a learner  grasp precisely the meaning  of terms? }



Motivated by these considerations and the goal of eventually develop an approach that could generalize to unseen words and even to the low resource languages scenario, we introduce the following hypernymy detection paradigm, \emph{\modelname}, where we augment distributional contextual models with that of learning terms representations from their definitions.  This paradigm has an important advantage in its \emph{powerful generalization}, as definitions are agnostic to specific domains and benchmarks, and are equally available for words regardless of their  frequency in a given data set. Consequently, the  task of identifying the relation between two terms is enhanced by the knowledge of the terms' definitions. Our model can be applied to any new terms in any domain, given some context of the term usage and their domain-agnostic definitions.
Moreover, given our learning approach -- we learn also the notion of {\em lexical entailment} between terms -- we can generalize to any lexical relation between terms.

Technically, we implement \modelname\enspace  by modifying the AttentiveConvNet \cite{DBLP00519}, a top-performing  system on a textual entailment benchmark \cite{DBLPBowmanAPM15}, to model the input ($x$, $d_x$; $y$, $d_y$), where $d_i$ ($i=x,y$) is the definition of term $i$. In contrast to earlier work which mostly built separate representations for terms $x$ and $y$, \modelname\enspace instead directly models the representation for each pair in  \{($x$, $y$), ($x$, $d_y$), ($d_x$, $y$), ($d_x$, $d_y$)\}, and then accumulates the four-way representations to form an overall representation for the input. 

In our experiments, we train \modelname\enspace on a task-agnostic annotated dataset, Wordnet, and test it on a broad array of open-domain hypernymy detection datasets. The results show the outstanding performance and strong generalization of  the \modelname\enspace model.

Overall, our contributions are as follows:
\begin{itemize}
[leftmargin=*]
\setlength\itemsep{0.01em}
\item To our knowledge, this is the first work in hypernymy detection that makes use of term definitions. Definitions provide complementary knowledge to distributional context, so that our model better tolerates unseen words, rare words and words with biased sense distribution.

\item \modelname\enspace  accounts for word sense when inferring the hypernymy relation.
This differs from much of the literature,  which usually derives sense-unaware representative vectors for terms -- earlier approaches would say `yes' if the relation holds for some combination of the terms' senses. 
\item \modelname\enspace has strong generalization capability -- once trained on a task-agnostic definition dataset, it can be used in different testbeds, and shows state-of-the-art results.
\end{itemize}

\section{Related Work}

The main novelty of our \modelname\enspace lies in the \emph{information resource} that is employed to represent the terms. Prior work in exploring information resources can be put into two categories: understanding terms by the co-occurring context in raw text, or grounding the terms in  open-domain objects.

\subsection{Mining Distributional Context from Text}

\paragraph{Window-based Context}
\newcite{DBLPBaroniBDS12} build distributional semantic vectors for terms from a concatenation of three corpora: the British National Corpus, WackyPedia and ukWac. Each entry in the vector is the PMI-formulated score from co-occurrence counts. Dimension reduction is conducted by Singular Value Decomposition (SVD) before feeding representation vectors to a classifier.


\paragraph{Dependency-based Context}
\newcite{DBLPRollerE16} compute a \emph{syntactic} distributional space for terms by counting their \emph{dependency neighbors} across the corpus.

\newcite{DBLPSantusSS17} further compare (i) contexts being parent and daughter nodes in the dependency tree, and (ii) contexts being the parent-sister pairs in the dependency tree.

\paragraph{Term Embeddings} Unspecialized term embeddings are not informative signals for detecting specific lexico-semantic relations. Hence, community often explicitly build transformation functions  from unspecialized embeddings to  relation-specialized embeddings.
\newcite{DBLPFuGQCWL14} first use the skip-gram model \cite{mikolov2013distributed} to learn generic term embeddings
from a large Chinese encyclopedia corpus, then  learn a projection function from the generic space to hypernymy space by annotated hypernymy pairs. Other work trying to specify the generic word embeddings to hypernymy detection task include \cite{DBLP06371,DBLPGlavasP17}.

Other advanced types of term embeddings specific to the hypernymy detection problem include Gaussian distributed embeddings \cite{DBLPVilnisM14}, non-negative embeddings \cite{DBLP00880}, magnitude-oriented embeddings \cite{DBLPNguyenKWV17}, and so on.






In our work, distributional context model is also applied. More specifically, we will directly use pretrained word embeddings as initial word representations and specialize them in training. In contrast, distributional context only acts as one side of information resource to express words, we focus on making use of a second side of information from word definitions to build a more robust system.

\subsection{Grounding  Terms to  Open-domain Objects}
\newcite{DBLPDoR12} build Wikipedia representations for input terms -- representing the input terms by a set of relevant Wikipedia pages.

\newcite{DBLPShwartzLDG15} represent each term pair as a set of paths which are extracted from different large-scale knowledge resources (DBPedia, Wikidata, Yago and WordNet), then train a classifier to determine whether the two terms satisfy a relation of interest given those path connections.

\newcite{DBLPYoungLHH14} map terms to a set of images, then determine the directional inference by conditional probability over statistic of image intersection.

Compared with mining of distributional context from text, these works switch the context from words to Wikipedia pages, KB paths or images. So, they share a similar mechanism while differing in the categories of entries in distributional vectors.

Our paradigm \modelname\enspace  shares the same inspiration with above distributional models. More importantly, It goes beyond the frame of distributional models by exploring a novel information resource -- definitions -- to derive the word semantics. 

\section{\modelname\enspace Model}

In this section, we first  give a brief review of a top-performing neural network for textual entailment -- AttentiveConvNet \cite{DBLP00519}, which acts as a base model to encode a pair of texts. Then, we elaborate on the adaptation we make towards AttentiveConvNet so that the resulting system can better serve the hypernymy detection problem. 


\subsection{AttentiveConvNet}\label{sec:abcnn}
AttentiveConvNet\footnote{https://github.com/yinwenpeng/Attentive\_Convolution} \cite{DBLP00519} is essentially a Siamese convolutional neural network (CNN) \cite{lecun1998gradient} equipped with an attention mechanism. It predicts the relationship of two sentences by accumulating the dominant features of  fine-grained alignments across sentences. The reason we base our system on this model is  two-fold: (i) AttentiveConvNet is one of the top-performing systems of modeling sentence pairs in textual entailment, and (ii) AttentiveConvNet implements the fine-grained cross-sentence alignments in the granularity of local windows; this makes it appropriate to reason between a definitional sentence and a term.

We use  bold uppercase, e.g.,
$\mathbf{H}$, for matrices;
bold lowercase,
e.g.,
$\mathbf{h}$,  for vectors;
bold lowercase with index,
e.g.,
$\mathbf{h}_i$,  for columns of
 $\mathbf{H}$; format $\mathbf{h}$[i] to denote the $i^{th}$ entry of vector $\mathbf{h}$; and non-bold lowercase for
scalars.

\begin{figure}
\centering
\includegraphics[width=0.45\textwidth]{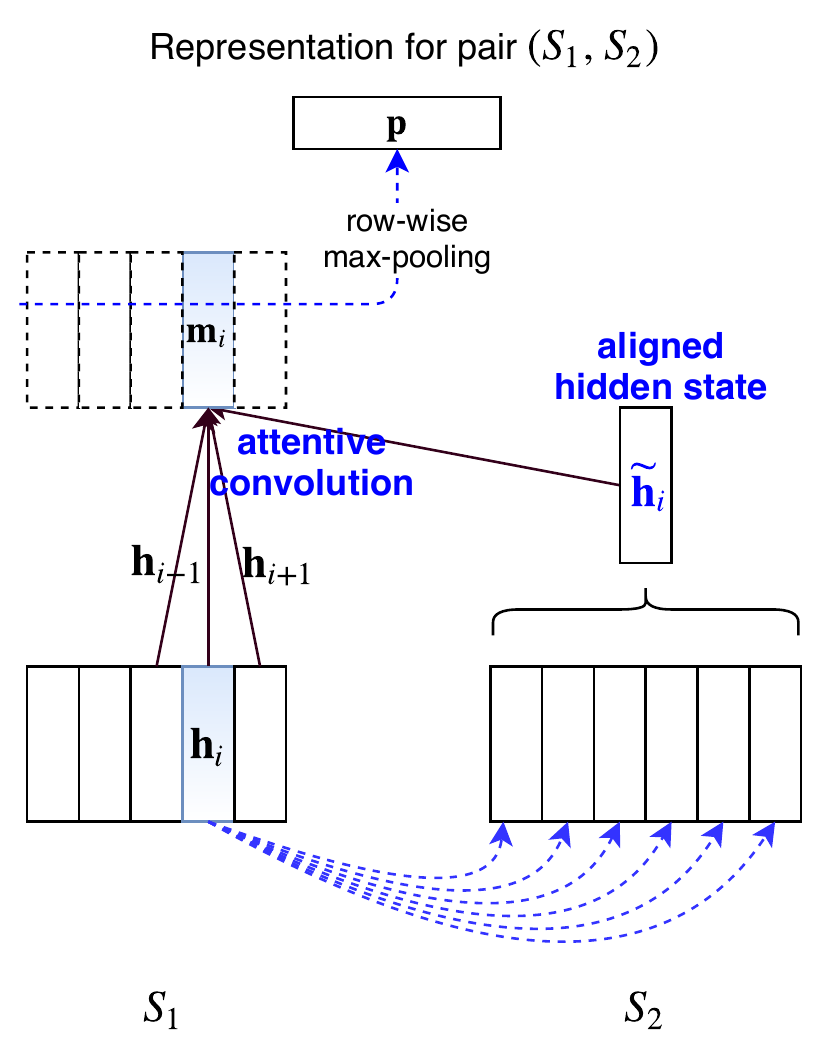}
\caption{{\small AttentiveConvNet models a sent. pair ($S_1$, $S_2$)}. In our work, $S_i$ ($i$=1,2) can be the definition sentence or the term itself (we treat a term as a short sentence.)}\label{fig:lightacnn}
\end{figure}

AttentiveConvNet, shown in Figure \ref{fig:lightacnn},
represents a
sentence $S$ ($S \in \{S_1, S_2\}$) of $n$ words as a sequence of
hidden states $\textbf{h}_i\in\mathbb{R}^d$
($i=1,2,\ldots, n$), forming a feature map
$\mathbf{H}\in\mathbb{R}^{d\times n}$, where $d$ is the
dimensionality of hidden states. Each  $\textbf{h}_i$ has a left context $\textbf{h}_{i-1}$ and a
right context $\textbf{h}_{i+1}$. Given feature maps $\mathbf{H}_1$ and  $\mathbf{H}_2$ for sentences $S_1$ and $S_2$ respectively, AttentiveConvNet  derives a representation for the pair ($S_1$, $S_2$). Unlike conventional CNNs over single sentences, AttentiveConvNet develops an attention mechanism to achieve fine-grained alignments automatically, then puts convolution filters over aligned hidden states together with their context.

Overall, AttentiveConvNet derives the pair representation in three steps. (i) A
matching  function 
determines how relevant each hidden
state in sentence $S_2$ is to the current
hidden state $\textbf{h}_i$ in sentence $S_1$. All hidden states in $S_2$ are then accumulated by weighted average to form an \emph{aligned hidden state} $\tilde{\mathbf{h}}_i$. (ii)
Convolution for position $i$ in $S_1$ integrates the  two aligned hidden
states ($\textbf{h}_i$, $\tilde{\textbf{h}}_i$) with 
context $\textbf{h}_{i-1}$ and $\textbf{h}_{i+1}$. (iii) Max-pooling over the generated group of hidden states in step (ii)  yields a representation for the pair ($S_1$, $S_2$). Next, we describe these processes in detail.

\textbf{Generation of Aligned Hidden States.} First, a
matching function $f_e(\textbf{h}_i, \textbf{h}^{S_2}_j)$ generates a  score $e_{i,j}$
 to evaluate how relevant  the two hidden states $\textbf{h}_i, \textbf{h}^{S_2}_j$ are.

Given the matching scores, the aligned hidden state $\tilde{\textbf{h}}_i$ in $S_2$  for hidden state $\mathbf{h}_i$ in $S_1$ is the weighted average of all hidden states in $S_2$:
\begin{equation}
\tilde{\textbf{h}}_i = \sum_j \mathrm{softmax}(\mathbf{e}_{i})[j]\cdot \mathbf{h}^{S_2}_j
\end{equation}
\enote{wp}{not clear $\mathbf{m}_i$ should be a vector or value in the vector}

\textbf{Attentive Convolution.} A position $i$ in $S_1$ has hidden state $\mathbf{h}_i$,  left
context $\mathbf{h}_{i-1}$, right context
$\mathbf{h}_{i+1}$ and aligned hidden state
$\tilde{\mathbf{h}}_i$ from $S_2$. Attentive convolution then  generates the higher-level representation for this combination:
\begin{align}\label{eq:aconv}
\mathbf{m}_i=\mathrm{tanh}(&\mathbf{W}\cdot [\mathbf{h}_{i-1},\mathbf{h}_i,\mathbf{h}_{i+1},\tilde{\mathbf{h}}_i]+\mathbf{b}) 
\end{align}
where parameters $\mathbf{W}\in\mathbb{R}^{d\times 4d}$, $\mathbf{b}\in\mathbb{R}^d$.

\textbf{Pair Representation Generation.} As Equation \ref{eq:aconv} shows, each $\mathbf{m}_i$ denotes the inference features between $\mathbf{h}_i$ and its alignment $\tilde{\mathbf{h}}_i$ in context. AttentiveConvNet  uses max-pooling over \{$\mathbf{m}_i$\} to get the overall representation $\mathbf{p}$ for the pair:
\begin{equation}
\textbf{p}[i] = \mathrm{max}(\mathbf{m}_{1}[i], \mathbf{m}_{2}[i], \cdots, \mathbf{m}_{n}[i])
\end{equation}

 Finally,  the representation $\mathbf{p}$ is used in classification. The whole model is learned in an end-to-end training\footnote{For more details, please refer to \cite{DBLP00519}.}.
\begin{figure}
\centering
\includegraphics[width=0.48\textwidth]{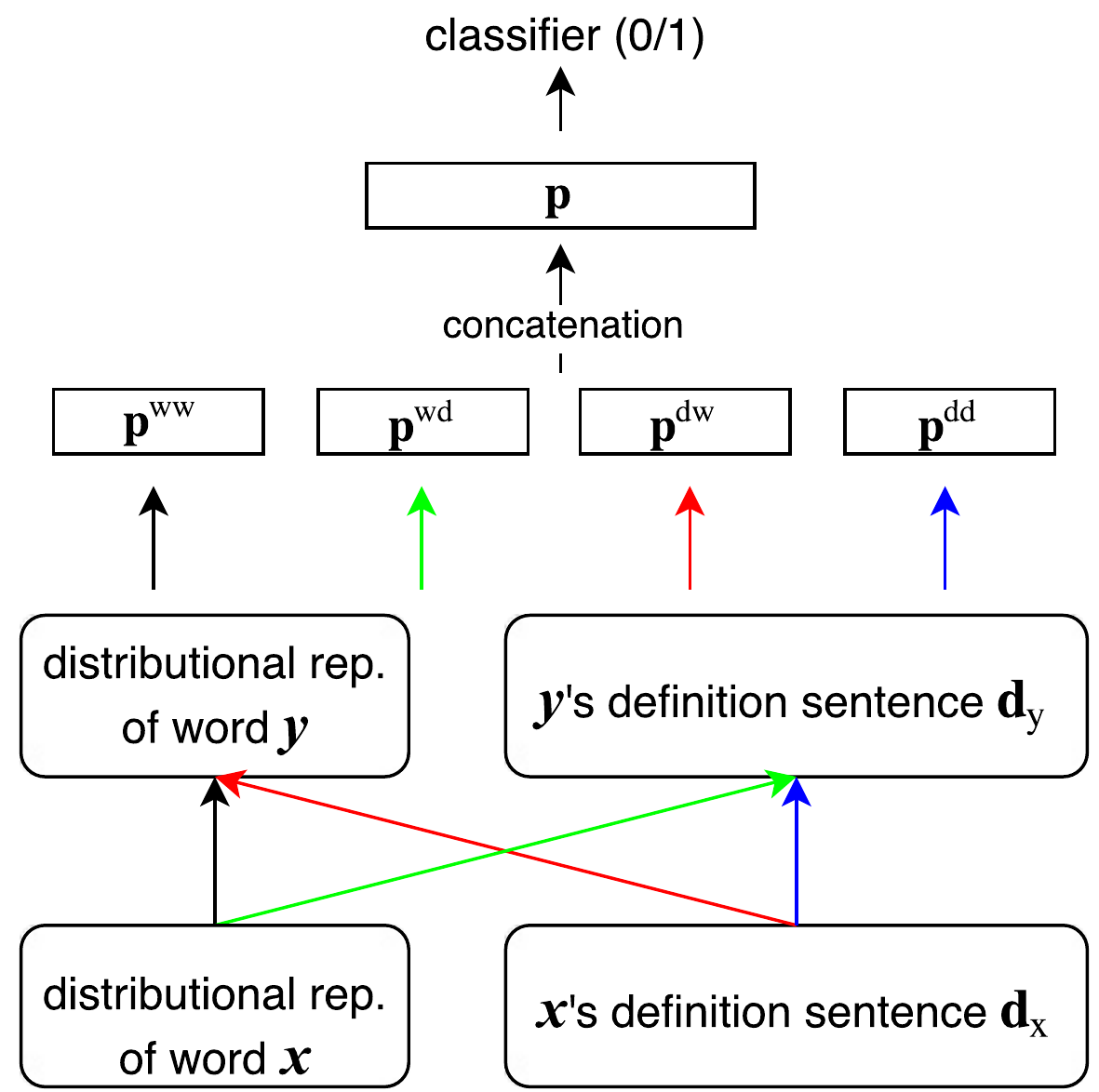}
\caption{\modelname\enspace -- combining distributional model with definition encoding}\label{fig:fourway}
\end{figure}

\begin{table*}
  \centering
 
  \begin{tabular}{l|c|cccc||c|cccc}
  & \multicolumn{5}{c||}{random split} & \multicolumn{5}{c}{ lexical split}\\
  & all &--$\mathbf{p}^{ww}$ & --$\mathbf{p}^{wd}$ & --$\mathbf{p}^{dw}$ & -- $\mathbf{p}^{dd}$ & all & --$\mathbf{p}^{ww}$ & --$\mathbf{p}^{wd}$ & -- $\mathbf{p}^{dw}$ & --$\mathbf{p}^{dd}$\\\hline
$\mathrm{\modelname}$ ($F_1$)&  .905& .874 & .896&.876&.881 & .887&.875 & .870 &.849& .862\\
$\mathrm{\modelname}$ (AP)& .933 & .902 & .921&.905&.909 & .900& .890 & .883&.880&.877\\\hline
w/o attention (F1) & \multicolumn{5}{c||}{.825} & \multicolumn{5}{c}{.743}\\
w/o definition (F1) & \multicolumn{5}{c||}{.734} & \multicolumn{5}{c}{.619}\\\hline
LSTM+atten. (F1) & \multicolumn{5}{c||}{.757} & \multicolumn{5}{c}{.685}
\end{tabular}
\caption{Tune HyperDef on $wn\_{dev}$}\label{tab:wndev}
\end{table*}
\subsection{Four-way AttentiveConvNet}
AttentiveConvNet originally works on sentence pairs. We  formulate the hypernymy detection problem as  $\{(x,d_x; y,d_y; 1/0)\}$.  Just as in \cite{DBLPShwartzGD16} which directly concatenates the term path representation vector with term embedding vectors as the classifier input, a simple combination of distributional models and definition encoding for us could be: separately learning  the distributional model over term embedding pairs and an AttentiveConvNet model over definition pairs, then concatenate their output representations. However, the analysis over dataset $\{(x,d_x; y,d_y; 1/0)\}$ hints that \modelname\enspace can obtain more indicative features by modeling (term, definition), which crosses the distributional models and definition encoding.  For example, the definition of term ``cat'' in WordNet is: \emph{feline mammal usually having thick soft fur and no ability to roar: domestic cats; wildcats}. Intuitively, when the system meets the pair (cat, mammal), it should be trivial to get the ``hypernymy'' decision since ``mammal'' appears in the definition sentence.

Inspired by this observation, we implement the  \modelname\enspace paradigm as four-way AttentiveConvNets, as Figure \ref{fig:fourway} shows, i.e., treating the two terms as word sequences as well, then do AttentiveConvNet over all four combinations: ($x$, $y$), ($x$, $d_y$), ($d_x$, $y$) and ($d_x$, $d_y$).

Assume we get four separate representations: $\mathbf{p}^{ww}$ from ($x$, $y$), $\mathbf{p}^{wd}$ from ($x$, $d_y$), $\mathbf{p}^{dw}$ from ($d_x$, $y$) and  $\mathbf{p}^{dd}$ from ($d_x$, $d_y$), as Section \ref{sec:abcnn} described. We construct the final representation for $(x,d_x; y,d_y)$ via concatenation:
\begin{equation}
\label{eq:conc}
\mathbf{p} = [\mathbf{p}^{ww},\mathbf{p}^{wd},\mathbf{p}^{dw},\mathbf{p}^{dd}]
\end{equation}
then $\mathbf{p}$ is fed to the final classifier.

AttentiveConvNet over ($x$, $y$) resembles the conventional hypernymy classifiers which take two representation vectors (one for $x$, the other for $y$) as input and output the label. Note that AttentiveConvNet puts filter weights over ($x$, $y$) to learn more abstract representations; this actually is in common with some literature such as \cite{DBLPFuGQCWL14,DBLP06371,DBLPGlavasP17}, which utilize weights to project  generic word representations into specified representations towards hypernymy annotations.

AttentiveConvNet over ($x$, $d_y$) and ($d_x$, $y$) compares a term with the descriptive sentence of the other term; this might provide  direct clues, as we discussed in the beginning of this subsection.

AttentiveConvNet over ($d_x$, $d_y$) resembles literature \cite{DBLPDoR12,DBLPYoungLHH14}.  \modelname\enspace provides an alternative resource for interpreting terms, resorting to definitional expressions  instead of Wikipedia pages or images.

Overall, our \modelname\enspace  combines strengths of (i) conventional supervised classifiers over context distributions, and (ii) rich interpretation of terms in broader knowledge bases.

\subsection{Analysis of \modelname}
Our \modelname\enspace has the following properties:
\begin{itemize}
\item \modelname\enspace combines distributional models with definition encoding, but it is not simply a concatenation of two independent subsystems. \modelname\enspace enables  modeling across (distributional context, definition). This is expected to generate more indicative features than a similar work \cite{DBLPShwartzGD16}, which simply concatenated distributional models with path-based models;
\item \modelname\enspace employs definitions to provide richer information for the terms. But it does not generate an auxiliary term representation vector from the definitive sentence as the literature \cite{DBLPHillCKB16} did. Instead, \modelname\enspace formulates a pair of input elements -- each can be a distributional vector or a definition representation -- into a cross-sentence attention mechanism, which directly yields a compact representation to the pair rather than two separate vectors for the two input elements. This is shown more effective to model the relationship of two pieces of text \cite{DBLP00519};
\item Distributional models and definitive sentences in \modelname\enspace provide complementary knowledge. For terms which can not retrieve a definition, \modelname\enspace still works -- just turning into a basic distributional model. This work uses WordNet and Wikipedia as example resources for the definition retrieval, more splendid resources will be developed  gradually in released \modelname\enspace models. We will also provide users the option to type into their definitions;
\item WordNet provides term definitions in the sense level, so the\modelname\enspace model is essentially trained in the sense level. For polysemy cases in testing, \modelname\enspace can simply test on all combinations of definitions, then pick the pair with the highest probability;
\item For terms that were never observed in training,  we expect context distributions, such as pretrained embeddings, and definitions are available, so \modelname\enspace is hardly influenced in this case. This is exactly the main advantage of \modelname: generalization. 
\end{itemize}

\section{Experiments}\label{sec:experiment}

\subsection{Pre-training of \modelname}
\paragraph{Dataset Preparation.}\label{sec:dataset}

As we aim to build a strongly generalizing hypernymy detector, 
the training data we collect here is expected to be task-agnostic. Hence, extracting from structured knowledge resources, such as WordNet \cite{DBLPMiller95}, Wikidata \cite{DBLPVrandecic12}, DBPedia \cite{DBLPAuerBKLCI07},
and Yago \cite{DBLPSuchanekKW07}, is preferred. Some literature, e.g., \cite{DBLPShwartzLDG15}, claim that there is limited coverage for almost all knowledge resources. For example, WordNet does not cover many proper-names (Donald Trump $\to$  president) or recent terminology (AlphaGo $\to$ computer program). Our data tends to alleviate this  challenge, since in testing,  descriptive sentences in  the \modelname\enspace paradigm can provide the precise and distinct features for terms even if these terms are OOV and in new types.

In this work, we  pick one of those knowledge resources -- WordNet -- to \emph{collect training data}. Specifically, our positive instances consist of (i)  all direct hypernymy pairs, and (ii) switched terms from the original hyponymy pairs.  Negative instances include (i) pairs with other relations such as antonym, synonym, and (ii) pairs of positive instances after exchanging the two terms.  Note that each term is accompanied by its definition \emph{in sense level}. So we get instances in form $(x,d_x; y,d_y; 1/0)$, where the binary value ``1/0'' indicates whether $y$ is $x$'s hypernymy or not.  Altogether, we collect about 900K instances with roughly a 8:1 ratio between negative and positive instances. 

In testing, we implement \modelname\enspace to retrieve definitions and distributional context  for terms automatically\footnote{In the released \modelname\enspace model, we will provide an option for users to input definitions.}. 

\paragraph{Random and Lexical Dataset Splits.}
In our primary dataset, we perform a \emph{random} split, with 80\% train, 10\% dev, and 10\% test. 

As pointed out by \newcite{DBLPLevyRBD15}, supervised distributional lexical inference methods tend to perform ``lexical memorization'', i.e., instead of learning a relation between the two terms, they mostly learn an independent property of term $y$ in the pair: whether $y$ is a ``prototypical hypernym'' or not. \newcite{DBLPLevyRBD15} suggest to splitting the train and test sets such that each will contain a distinct vocabulary (``\emph{lexical split}''), in order to prevent the model from overfitting by lexical memorization.

In the current phase, we use notations \emph{wn\_train}, \emph{wn\_dev}, and \emph{wn\_test} to refer to the three parts. Note that \emph{wn\_train} and \emph{wn\_dev} will be used to train  and tune the \modelname\enspace model, while \emph{wn\_test} is set to show how well the model performs in WordNet domain -- it is not expected to act as a testbed in real benchmarks. In experiments, we will compare our model in \emph{random} and \emph{lexical} splits.

\paragraph{Training Setup.} Given  \emph{wn\_train} in form \{($x$, $d_x$; $y$, $d_y$; 1/0)\}, a binary classifier via logistic regression is trained over the pair representation $\mathbf{p}$ obtained from Equation \ref{eq:conc}, predicting \emph{1} or \emph{0} for the hypernymy relation. The objective function is implemented through negative log-likelihood. 
 Terms and words in definitions are initialized by 300d Word2Vec embeddings \cite{mikolov2013distributed} and kept unchanged in training. This benefits the generalization as it ensures that the words  in training and the new words in test data lie in the same space. All hidden sizes are 300 as well.  The whole system is trained by AdaGrad \cite{duchi2011adaptive} with initial learning rate 0.02.

We first run the   \modelname\enspace in \emph{wn\_test} to check if it is effective in the WordNet domain. Then we test it in some open-domain benchmarks.  \emph{Note that all experiments use the   \modelname\enspace models pretrained over \emph{wn\_train}.}
\subsection{Performance within WordNet}
As mentioned in Section \ref{sec:dataset},  \emph{wn\_train}, \emph{wn\_dev} and \emph{wn\_test} have two distinct setups: ``random split'' and ``lexical split'', inspired by the ``lexical memorization'' observation \cite{DBLPLevyRBD15}.

We first tune the parameters in $wn\_{train}$ and search the best system layout based on $wn\_{dev}$. $F_1$ and average precision (AP) are reported. 
Table \ref{tab:wndev} lists the performance records, with the first block for ``random split'' and the second block for ``lexical split''.

We  first discuss three baselines: (i) ``w/o definition'': We discard definitions and only use distributional model, i.e.,  a logistic regression classifier (LR) over the concatenated ($x$, $y$) embeddings from Word2Vec. Its performance drops 11.5\% from ``random split'' to ``lexical split''. This is within expectation as \newcite{DBLPLevyRBD15} concluded that this baseline is not effective in learning genuine term relations; (ii) ``w/o attention'': We discard the attention mechanism in AttentiveConvNet, resulting in a bi-CNN structure. It works on instances \{($x$, $d_x$; $y$, $d_y$; 1/0)\},  a vanilla CNN  is used to encode the definition sentence into a dense representation vector. So, each term in ($x$, $y$) will get two separate representation vectors (one is from Word2Vec, the other from the definition); finally totally four representation vectors are concatenated and fed to the LR. This baseline works much better than ``w/o definition'' (improvements of 9\% $\sim$ 11\%). Their comparison shows that incorporating term definitions in reasoning process is promising; (iii) ``LSTM+attention'' \cite{entail2016}. A representative attention mechanism in LSTM \cite{hochreiter1997long} for textual entailment. We apply it in the same way as our four-way AttentiveConvNet, however, found it performs poorly. We suspect that this is due to two reasons: i) Though there is entailment or hypernymy relation between a term pair, e.g., (``cat'', ``animal''), unfortunately there is no clear clue of that relation between their definition pair \emph{if considering all the information contained in the definitions}. For example, ``cat'' -- ``\emph{a small domesticated carnivorous mammal with soft fur, a short snout, and retractile claws. It is widely kept as a pet or for catching mice, and many breeds have been developed}'', and ``animal'' -- ``\emph{a living organism that feeds on organic matter, typically having specialized sense organs and nervous system and able to respond rapidly to stimuli}''. Apparently, we can not infer the whole definition of ``animal'' by cat's definition. Instead, their help  mainly comes from some key-phrases, such as ``domesticated carnivorous mammal'', ``living organism'' and so on. LSTM, encoding the whole word sequences in attention, potentially would be misled. Our approach relies on convolution filters and max-pooling, excelling in modeling keywords-driving features \cite{DBLP0001KYS17}. This baseline indicates the overall strength of our system comes from the definition incorporation as well as an appropriate encoder.

Considering the whole table, we observe that: (i) \modelname\enspace models have  pretty close performances in ``random split'' and ``lexical split'' -- mostly within 2$\sim$3\%. This strongly indicates that \modelname\enspace is less influenced by the ``lexical memorization'' problem. Our systems, equipped with definition encoding, show promising generalization (at least in WordNet  domain); (ii) Though \modelname\enspace models in ``all'' setup behave similarly in random split and lexical split, the detailed contributions of $\mathbf{p^{ww}}$, $\mathbf{p^{wd}}$, $\mathbf{p^{dw}}$ and $\mathbf{p^{dd}}$ differ in the two settings. To be specific, in ``$wn\_{dev}$ (random split)'', there is no clear winner among \{$\mathbf{p}^{ww}$, $\mathbf{p}^{dw}$, $\mathbf{p}^{dd}$\}, $\mathbf{p}^{wd}$ contributes consistently less than the other three. In ``$wn\_{dev}$ (lexical split)'', instead, $\mathbf{p}^{wd}$, $\mathbf{p}^{dw}$ and $\mathbf{p}^{dd}$ perform similarly while $\mathbf{p}^{ww}$ performs worst.  This indicates that when dealing with unseen terms, definition-based components in \modelname\enspace play a dominant role.

\begin{table}
 \setlength{\abovecaptionskip}{5pt}
  \centering
  \begin{tabular}{l|cc}
&  random  &  lexical \\\hline
$F_1$& .902 & .881\\
AP&.915 & .891\\
\end{tabular}
\caption{Pretrained HyperDef  on $wn\_{test}$}\label{tab:wntest}
\end{table}

Experiments on $wn\_{dev}$ enable  to store the best  \modelname\enspace models -- concatenation over the four representations: $\mathbf{p^{ww}}$, $\mathbf{p^{wd}}$, $\mathbf{p^{dw}}$ and $\mathbf{p^{dd}}$. Then we reload the pretrained models and report their performance on $wn\_{test}$, as shown in Table \ref{tab:wntest}. From  Table \ref{tab:wndev} to Table \ref{tab:wntest}, we observe pretty small drop in performance -- mostly $\sim$ 1\% . This preliminarily demonstrates the strong generalization. 

Next, we test the best  \modelname\enspace models pretrained on ``$wn\_{train}$ (lexical split)'' in open domain benchmarks.

\subsection{Performance in Open-domain Datasets}
First, we use four widely-explored datasets: \textbf{BLESS} \cite{baroni2011we}, \textbf{EVALution} \cite{santus2015evalution}, \textbf{Lenci/Benotto} \cite{benotto2015distributional}, and \textbf{Weeds} \cite{DBLPWeedsCRWK14}. They were constructed either using knowledge resources (e.g. WordNet, Wikipedia), crowd-sourcing or both. The instance sizes of hypernymy and ``other'' relation types  are detailed in Table \ref{tab:4benchmark}. We also report ``\#OOV\_pair'', the proportions of unseen term pairs in above four datasets regarding  the training set of \modelname, i.e., \emph{wn\_train} in Section \ref{sec:dataset}. We notice that most term pairs in BLESS and Lenci/Benotto datasets are unseen in wn\_train.

\begin{table}
  \centering
  \begin{tabular}{l|ccc}
  dataset &  \#hyper.  & \#others  & \#OOV\_pair\\\hline
  BLESS & 1.337 & 25,217 & 99.04\%\\
EVALution &  3,637 & 9,828 & 78.86\%\\
{\small Lenci/Benotto} &  1,933 & 3,077 & 92.50\%\\
Weeds &  1,469 & 1,459 & 71.54\%

\end{tabular}
\caption{Statistics of four benchmarks. ``\#OOV\_pair'': the proportions of unseen term pairs regarding  the training set (i.e., \emph{wn\_train} in Section \ref{sec:dataset}) of \modelname.}\label{tab:4benchmark}
\end{table}
\begin{table*}
\setlength{\tabcolsep}{2.5pt}
  \centering
 
  \begin{tabular}{l|cc|cc|cc|cc}
   & \multicolumn{2}{c}{BLESS} & \multicolumn{2}{c}{EVALuation}& \multicolumn{2}{c}{Benotto}& \multicolumn{2}{c}{Weeds}\\
  Model & AP & AP@100& AP & AP@100& AP & AP@100& AP & AP@100\\\hline
  Best-Unsuper \cite{DBLPSantusSS17} & .051 & .540 & .353 & .661 & .382 & .617 & .441 & .911\\
  Concat-SVM {\small\cite{DBLPGlavasP17}} & .097 & .235 & .321 & .329 & .523 & .586 & .644 & .793\\
  DUAL-T \cite{DBLPGlavasP17} & .487 & .823 & .446 & .866 & .557 & .847 & .774 & .985\\
  HyperScore \cite{DBLPNguyenKWV17} & .454 & -- & .538 & -- & .574 & -- & .850 & --\\
  \hline
  $\mathrm{\modelname}_{TopDef}$ &\textbf{.595} & .749 & .524 & \textbf{.867} & .557 & .825 & \textbf{.872}& \textbf{.989}\\

  $\mathrm{\modelname}_{AllDef}$ &\textbf{.508} & \textbf{.872} & \textbf{.623} & \textbf{.927} & \textbf{.576} & \textbf{.909} &\textbf{.889} & \textbf{.991}

\end{tabular}
\caption{System comparison on BLESS, EVALution, Benotto and Weeds datasets}\label{tab:4benchmarkresult}
\end{table*}

First, we extract the term's \emph{all sense definitions} from WordNet based on  the term string.  For a few instances which contain terms not covered by WordNet, such as proper noun ``you'', ``everybody'' etc, we set definitions the same as the term strings (this preprocessing does not influence results, just for making the system uniformed).  Then, we apply the pre-trained $\mathrm{\modelname}$ model \emph{on the test sets of the four benchmarks},   discriminating hypernymy from ``other'' relations. AP and AP@100 are reported. As WordNet sorts sense definitions by sense frequency \cite{DBLPMiller95}, we test the term pairs in two  ways: (i) Only choose the top-1 sense definition to denote a term, reported as ``$\mathrm{\modelname}_{TopDef}$''; (ii) Keep all sense definitions for those terms, then test on all sense combinations and pick the highest probability as the term pair score, reported as``$\mathrm{\modelname}_{AllDef}$''.

We compare $\mathrm{\modelname}$ with baselines: (i) \textbf{Best-Unsuper}. The best unsupervised method in \cite{DBLPSantusSS17}, implemented by similarity measurement over weighted dependency-based context vectors; (ii) \textbf{Concat-SVM} \cite{DBLPGlavasP17}. An SVM model with RBF kernel is trained on concatenation of unspecialized concept embeddings \cite{DBLParoniBDS12}; (iii) \textbf{DUAL-T} \cite{DBLPGlavasP17}. Using dual tensors, DUAL-T transforms unspecialized embeddings into asymmetrically specialized representations -- sets of specialized vectors -- which are next used to predict whether the asymmetric relation holds between the concepts; (iv) \textbf{HyperScore} \cite{DBLPNguyenKWV17}. The state-of-the-art system. It uses a large-scale hypernymy pair set to guide the learning of hierarchical word embeddings in hypernymy-structured space.

Table \ref{tab:4benchmarkresult} clearly demonstrates the superiority of our $\mathrm{\modelname}$ models over other systems. The three baselines Concat-SVM, DUAL-T and HyperScore are more in line with $\mathrm{\modelname}$ since they did supervised learning over  large numbers of annotated pairs. $\mathrm{\modelname}$ integrates term definitions, which is shown effective in improving the performance across different testbeds.

In addition, \modelname$_{AllDef}$ consistently outperforms \modelname$_{TopDef}$. This makes sense as \modelname$_{TopDef}$ may be misled by  incorrect definitions. In addition, the superiority of \modelname$_{AllDef}$ clearly supports the effectiveness of \modelname\enspace in dealing with polysemy cases.

Above four benchmarks  are relatively small and contain  common  words mostly. In real-world applications,  there is a need to figure out the hypernymy relation between common nouns and proper nouns \cite{DBLPDoR12}. Taking ``(Champlin, city)'' for example, ``Champlin''  is not covered by WordNet vocabulary, thus uncovered by $wn\_{train}$ -- the training data of our $\mathrm{\modelname}$ model. Motivated, we further test $\mathrm{\modelname}$ on the following dataset.

\textbf{HypeNet Dataset}. \newcite{DBLPShwartzGD16} construct this dataset  by extracting hypernymy relations from several resources: WordNet, DBPedia, Wikidata and Yago.  Like our collected data, term pairs in other relations are considered as negative instances. It maintains a ratio of 1:4 positive to negative pairs.

Similarly, HypeNet dataset has  ``random split'' and 	``lexical split'' as well; their sizes are list in Table \ref{tab:hypenetdata}. HypeNet contains lots of locations, e.g., (Champlin, city), and  organizations, e.g., (Telegram, company) and (Sheetz, company). We first try to extract definitions for those terms from WordNet, if fail, then we extract from Wikipedia pages, treating the first sentence as a definition. 

We play \modelname\enspace in two different ways, one testing its ``\emph{pre-trained}'' model on HypeNet's test data, the other -- ``\emph{specialized}'' -- training \modelname\enspace on HypeNet's training data then test  on HypeNet's test data like other baselines did.

\begin{table}
\setlength{\tabcolsep}{3pt}
  \centering
 
  \begin{tabular}{l|cccc}
Dataset & train & dev & test &  \#OOV\_pair\\\hline
{\small HypeNet (rnd)} & 49.5K  & 3.5K  & 17.7K & 95.56\%\\
{\small HypeNet (lex)} & 20.3K  & 1.4K  & 6.6K & 95.33\%

\end{tabular}
\caption{Statistics of HypeNet dataset. ``\#OOV\_pair'' is for ``test'' regarding the ``wn\_train'' of \modelname.}\label{tab:hypenetdata}
\end{table}

\begin{table}
\setlength{\tabcolsep}{3pt}
  \centering
 
  \begin{tabular}{l|ccc|ccc}
  
  & \multicolumn{3}{c|}{Lex. split} & \multicolumn{3}{c}{Rand. split}\\
  Model & P & R & $F_1$ & P & R & $F_1$\\\hline
  HypeNet & .809 &.617 & .700 &.913 &.890 &.901\\
  DUAL-T & .705 & .785 &.743 & .933 & .826 & .876\\\hline

pre-trained & .572&.717&.637&.474&.601&.530\\

    specialized &.670 &.914&\textbf{.773}& .892& .935&\textbf{.913}\\

\end{tabular}
\caption{System comparison on HypeNet test}\label{tab:hypenetresult}
\end{table}

Table \ref{tab:hypenetresult} shows: 
(i) If trained on the specific training data of HypeNet, our system \modelname\enspace can get state of the art performance. This indicates the superiority of our model over baseline systems.

(i) Our pretrained $\mathrm{\modelname}$ model performs less satisfactorily. Only the result on ``Lex. split'' is relatively close to the outstanding baselines. This makes sense as baseline systems are specified by the HypeNet training set while our pretrained model comes from a different domain. We studied the dataset and found following problems.

\paragraph{Error Analysis.} Two error sources are observed. (i) Wrong definition. For example, the system obtains the definition ``\emph{a substance or treatment with no active therapeutic effect}'' for the term ``Placebo'' in the pair (Placebo, song); however, a successful detection requires mining another definition -- ``\emph{are an alternative rock band, formed in London, England in 1994 by singer-guitarist Brian Molko and guitarist-bassist Stefan Olsdal}'' which depicts the article title ``Placebo (band)''. This is a common problem due to the ambiguity of entity mentions. To relieve this, we plan to refine the definition retrieval by more advanced entity linking techniques, or retrieve all highly related definitions and test as in polysemy cases (recall that in Table \ref{tab:4benchmarkresult} we showed \modelname\enspace has more robust performance while addressing polysemy terms);  (ii) Misleading information in definitions. Our system  predicts ``1'' for the pair (Aurangabad, India); we analyze the definition of ``Aurangabad'': \emph{ is a city in the Aurangabad district of Maharashtra state in India}. We intentionally removed the phrase ``in India'', and then the system predicts ``0''. This demonstrates that definitions indeed provide informative knowledge about terms, but a system must be intelligent to avoid being misled; (iii) We miss a common embedding space to initialize single words and (multi-word) entities. To generalize well to new entities, the model has to presume the  new entities and the known terms lie in the same representation space. However, most  pretrained embedding sets cover pretty limited entities. To learn uniformed word and entity embeddings, we may need to combine unstructured text corpus, semi-structured data (e.g., Wikipedia) and structured knowledge bases together.  We will advance this data preprocessing component -- the access of term definitions and term representations -- in our released system.

\section{Conclusion}
In this work, we introduced a novel approach to detecting hypernymy relations by incorporating term definitions. We extracted a task-agnostic annotated data from WordNet, then trained a neural network  to generate a universal hypernymy detector, \modelname. $\mathrm{\modelname}$, once trained,  performs competitively  in diverse open-domain benchmarks, even though it was not fine-tuned on those benchmark-specific training sets. This validates the powerful generalization of our model \modelname.
Our hope, and one of the key future directions following this work is to generalize this approach to the low-resource language setting.  

\subsection*{Acknowledgments}
We thank all the reviewers for providing insightful comments and critiques. This research is supported in part by DARPA under agreement number FA8750-13-2-0008, and by a gift from Google. 

\bibliography{ccg-compact,naaclhlt2018}

\begin{thebibliography}{}
\expandafter\ifx\csname natexlab\endcsname\relax\def\natexlab#1{#1}\fi

\bibitem[{Auer et~al.(2007)Auer, Bizer, Kobilarov, Lehmann, Cyganiak, and
  Ives}]{DBLPAuerBKLCI07}
S{\"{o}}ren Auer, Christian Bizer, Georgi Kobilarov, Jens Lehmann, Richard
  Cyganiak, and Zachary~G. Ives. 2007.
\newblock Dbpedia: {A} nucleus for a web of open data.
\newblock In {\em Proceedings of {ISWC}/{ASWC}\/}. pages 722--735.

\bibitem[{Baroni et~al.(2012{\natexlab{a}})Baroni, Bernardi, Do, and
  Shan}]{DBLParoniBDS12}
Marco Baroni, Raffaella Bernardi, Ngoc{-}Quynh Do, and Chung{-}chieh Shan.
  2012{\natexlab{a}}.
\newblock Entailment above the word level in distributional semantics.
\newblock In {\em Proceedings of {EACL}\/}. pages 23--32.

\bibitem[{Baroni et~al.(2012{\natexlab{b}})Baroni, Bernardi, Do, and
  Shan}]{DBLPBaroniBDS12}
Marco Baroni, Raffaella Bernardi, Ngoc{-}Quynh Do, and Chung{-}chieh Shan.
  2012{\natexlab{b}}.
\newblock Entailment above the word level in distributional semantics.
\newblock In {\em Proceedings of {EACL}\/}. pages 23--32.

\bibitem[{Baroni and Lenci(2011)}]{baroni2011we}
Marco Baroni and Alessandro Lenci. 2011.
\newblock How we {BLESSed} distributional semantic evaluation.
\newblock In {\em Proceedings of the GEMS 2011 Workshop on GEometrical Models
  of Natural Language Semantics\/}. pages 1--10.

\bibitem[{Benotto(2015)}]{benotto2015distributional}
Giulia Benotto. 2015.
\newblock Distributional models for semantic relations: A study on hyponymy and
  antonymy.
\newblock {\em PhD Thesis, University of Pisa\/} .

\bibitem[{Bowman et~al.(2015)Bowman, Angeli, Potts, and
  Manning}]{DBLPBowmanAPM15}
Samuel~R. Bowman, Gabor Angeli, Christopher Potts, and Christopher~D. Manning.
  2015.
\newblock A large annotated corpus for learning natural language inference.
\newblock In {\em Proceedings of {EMNLP}\/}. pages 632--642.

\bibitem[{Chang et~al.(2017)Chang, Wang, Vilnis, and McCallum}]{DBLP00880}
Haw{-}Shiuan Chang, ZiYun Wang, Luke Vilnis, and Andrew McCallum. 2017.
\newblock Unsupervised hypernym detection by distributional inclusion vector
  embedding.
\newblock {\em CoRR\/} abs/1710.00880.

\bibitem[{Dagan et~al.(2013)Dagan, Roth, Sammons, and Zanzoto}]{DRSZ13}
Ido Dagan, Dan Roth, Mark Sammons, and Fabio~Massimo Zanzoto. 2013.
\newblock Recognizing textual entailment: Models and applications.

\bibitem[{Do and Roth(2012)}]{DBLPDoR12}
Quang~Xuan Do and Dan Roth. 2012.
\newblock Exploiting the wikipedia structure in local and global classification
  of taxonomic relations.
\newblock {\em Natural Language Engineering\/} 18(2):235--262.

\bibitem[{Duchi et~al.(2011)Duchi, Hazan, and Singer}]{duchi2011adaptive}
John Duchi, Elad Hazan, and Yoram Singer. 2011.
\newblock Adaptive subgradient methods for online learning and stochastic
  optimization.
\newblock {\em JMLR\/} 12:2121--2159.

\bibitem[{Fellbaum(1998)}]{DBLPMiller95}
Christiane Fellbaum. 1998.
\newblock {\em WordNet: An Electronic Lexical Database\/}.
\newblock MIT Press.

\bibitem[{Fu et~al.(2014)Fu, Guo, Qin, Che, Wang, and Liu}]{DBLPFuGQCWL14}
Ruiji Fu, Jiang Guo, Bing Qin, Wanxiang Che, Haifeng Wang, and Ting Liu. 2014.
\newblock Learning semantic hierarchies via word embeddings.
\newblock In {\em Proceedings of {ACL}\/}. pages 1199--1209.

\bibitem[{Glavas and Ponzetto(2017)}]{DBLPGlavasP17}
Goran Glavas and Simone~Paolo Ponzetto. 2017.
\newblock Dual tensor model for detecting asymmetric lexico-semantic relations.
\newblock In {\em Proceedings of {EMNLP}\/}. pages 1757--1767.

\bibitem[{Hearst(1992)}]{DBLPHearst92}
Marti~A. Hearst. 1992.
\newblock Automatic acquisition of hyponyms from large text corpora.
\newblock In {\em Proceedings of {COLING}\/}. pages 539--545.

\bibitem[{Hill et~al.(2016)Hill, Cho, Korhonen, and Bengio}]{DBLPHillCKB16}
Felix Hill, KyungHyun Cho, Anna Korhonen, and Yoshua Bengio. 2016.
\newblock Learning to understand phrases by embedding the dictionary.
\newblock {\em {TACL}\/} 4:17--30.

\bibitem[{Hochreiter and Schmidhuber(1997)}]{hochreiter1997long}
Sepp Hochreiter and J{\"u}rgen Schmidhuber. 1997.
\newblock Long short-term memory.
\newblock {\em Neural computation\/} 9(8):1735--1780.

\bibitem[{LeCun et~al.(1998)LeCun, Bottou, Bengio, and
  Haffner}]{lecun1998gradient}
Yann LeCun, L{\'e}on Bottou, Yoshua Bengio, and Patrick Haffner. 1998.
\newblock Gradient-based learning applied to document recognition.
\newblock In {\em Proceedings of the IEEE\/}. pages 2278--2324.

\bibitem[{Levy et~al.(2015)Levy, Remus, Biemann, and Dagan}]{DBLPLevyRBD15}
Omer Levy, Steffen Remus, Chris Biemann, and Ido Dagan. 2015.
\newblock Do supervised distributional methods really learn lexical inference
  relations?
\newblock In {\em Proceedings of {NAACL}\/}. pages 970--976.

\bibitem[{Mikolov et~al.(2013)Mikolov, Sutskever, Chen, Corrado, and
  Dean}]{mikolov2013distributed}
Tomas Mikolov, Ilya Sutskever, Kai Chen, Greg~S Corrado, and Jeff Dean. 2013.
\newblock Distributed representations of words and phrases and their
  compositionality.
\newblock In {\em Proceedings of NIPS\/}. pages 3111--3119.

\bibitem[{Mintz et~al.(2009)Mintz, Bills, Snow, and Jurafsky}]{DBLPMintzBSJ09}
Mike Mintz, Steven Bills, Rion Snow, and Daniel Jurafsky. 2009.
\newblock Distant supervision for relation extraction without labeled data.
\newblock In {\em Proceedings of {ACL}\/}. pages 1003--1011.

\bibitem[{Nguyen et~al.(2017)Nguyen, K{\"{o}}per, {Schulte im Walde}, and
  Vu}]{DBLPNguyenKWV17}
Kim~Anh Nguyen, Maximilian K{\"{o}}per, Sabine {Schulte im Walde}, and
  Ngoc~Thang Vu. 2017.
\newblock Hierarchical embeddings for hypernymy detection and directionality.
\newblock In {\em Proceedings of {EMNLP}\/}. pages 233--243.

\bibitem[{Rockt{\"a}schel et~al.(2016)Rockt{\"a}schel, Grefenstette, Hermann,
  Ko{\v{c}}isk{\`y}, and Blunsom}]{entail2016}
Tim Rockt{\"a}schel, Edward Grefenstette, Karl~Moritz Hermann, Tom{\'a}{\v{s}}
  Ko{\v{c}}isk{\`y}, and Phil Blunsom. 2016.
\newblock Reasoning about entailment with neural attention.
\newblock In {\em Proceedings of ICLR\/}.

\bibitem[{Roller and Erk(2016)}]{DBLPRollerE16}
Stephen Roller and Katrin Erk. 2016.
\newblock Relations such as hypernymy: Identifying and exploiting hearst
  patterns in distributional vectors for lexical entailment.
\newblock In {\em Proceedings of {EMNLP}\/}. pages 2163--2172.

\bibitem[{Roller et~al.(2014)Roller, Erk, and Boleda}]{DBLPRollerEB14}
Stephen Roller, Katrin Erk, and Gemma Boleda. 2014.
\newblock Inclusive yet selective: Supervised distributional hypernymy
  detection.
\newblock In {\em Proceedings of {COLING}\/}. pages 1025--1036.

\bibitem[{Santus et~al.(2015)Santus, Yung, Lenci, and
  Huang}]{santus2015evalution}
Enrico Santus, Frances Yung, Alessandro Lenci, and Chu-Ren Huang. 2015.
\newblock {EVALution} 1.0: {An} evolving semantic dataset for training and
  evaluation of distributional semantic models.
\newblock In {\em Proceedings of the 4th Workshop on Linked Data in
  Linguistics\/}. pages 64--69.

\bibitem[{Saxena et~al.(2007)Saxena, Sambhu, Kaushik, and
  Subramaniam}]{DBLPSaxenaSKS07}
Ashish~Kumar Saxena, Ganesh~Viswanath Sambhu, Saroj Kaushik, and L.~Venkata
  Subramaniam. 2007.
\newblock {IITD-IBMIRL} system for question answering using pattern matching,
  semantic type and semantic category recognition.
\newblock In {\em Proceedings of {TREC}\/}.

\bibitem[{Shwartz et~al.(2016)Shwartz, Goldberg, and Dagan}]{DBLPShwartzGD16}
Vered Shwartz, Yoav Goldberg, and Ido Dagan. 2016.
\newblock Improving hypernymy detection with an integrated path-based and
  distributional method.
\newblock In {\em Proceedings of {ACL}\/}.

\bibitem[{Shwartz et~al.(2015)Shwartz, Levy, Dagan, and
  Goldberger}]{DBLPShwartzLDG15}
Vered Shwartz, Omer Levy, Ido Dagan, and Jacob Goldberger. 2015.
\newblock Learning to exploit structured resources for lexical inference.
\newblock In {\em Proceedings of {CoNLL}\/}. pages 175--184.

\bibitem[{Shwartz et~al.(2017)Shwartz, Santus, and
  Schlechtweg}]{DBLPSantusSS17}
Vered Shwartz, Enrico Santus, and Dominik Schlechtweg. 2017.
\newblock Hypernyms under siege: Linguistically-motivated artillery for
  hypernymy detection.
\newblock In {\em Proceedings of {EACL}\/}. pages 65--75.

\bibitem[{Snow et~al.(2004)Snow, Jurafsky, and Ng}]{DBLPSnowJN04}
Rion Snow, Daniel Jurafsky, and Andrew~Y. Ng. 2004.
\newblock Learning syntactic patterns for automatic hypernym discovery.
\newblock In {\em Proceedings of {NIPS}\/}. pages 1297--1304.

\bibitem[{Suchanek et~al.(2007)Suchanek, Kasneci, and
  Weikum}]{DBLPSuchanekKW07}
Fabian~M. Suchanek, Gjergji Kasneci, and Gerhard Weikum. 2007.
\newblock Yago: {A} core of semantic knowledge.
\newblock In {\em Proceedings of {WWW}\/}. pages 697--706.

\bibitem[{Vilnis and McCallum(2015)}]{DBLPVilnisM14}
Luke Vilnis and Andrew McCallum. 2015.
\newblock Word representations via gaussian embedding.
\newblock In {\em Proceedings of {ICLR}\/}.

\bibitem[{Vrandecic(2012)}]{DBLPVrandecic12}
Denny Vrandecic. 2012.
\newblock Wikidata: {A} new platform for collaborative data collection.
\newblock In {\em Proceedings of {WWW}\/}. pages 1063--1064.

\bibitem[{Vulic and Mrksic(2017)}]{DBLP06371}
Ivan Vulic and Nikola Mrksic. 2017.
\newblock Specialising word vectors for lexical entailment.
\newblock {\em CoRR\/} abs/1710.06371.

\bibitem[{Weeds et~al.(2014)Weeds, Clarke, Reffin, Weir, and
  Keller}]{DBLPWeedsCRWK14}
Julie Weeds, Daoud Clarke, Jeremy Reffin, David~J. Weir, and Bill Keller. 2014.
\newblock Learning to distinguish hypernyms and co-hyponyms.
\newblock In {\em Proceedings of {COLING}\/}. pages 2249--2259.

\bibitem[{Yin et~al.(2017)Yin, Kann, Yu, and Sch{\"{u}}tze}]{DBLP0001KYS17}
Wenpeng Yin, Katharina Kann, Mo~Yu, and Hinrich Sch{\"{u}}tze. 2017.
\newblock Comparative study of {CNN} and {RNN} for natural language processing.
\newblock {\em CoRR\/} abs/1702.01923.

\bibitem[{Yin and Sch{\"{u}}tze(2017)}]{DBLP00519}
Wenpeng Yin and Hinrich Sch{\"{u}}tze. 2017.
\newblock Attentive convolution.
\newblock {\em CoRR\/} abs/1710.00519.

\bibitem[{Young et~al.(2014)Young, Lai, Hodosh, and
  Hockenmaier}]{DBLPYoungLHH14}
Peter Young, Alice Lai, Micah Hodosh, and Julia Hockenmaier. 2014.
\newblock From image descriptions to visual denotations: New similarity metrics
  for semantic inference over event descriptions.
\newblock {\em {TACL}\/} 2:67--78.

\end{thebibliography}
\bibliographystyle{acl_natbib}
\end{document}